%% file: ms.tex
\documentclass[conference]{IEEEtran}
\pdfoutput=1
\IEEEoverridecommandlockouts
\usepackage{cite}
\usepackage{amsmath,amssymb,amsfonts}
\usepackage{algorithmic}
\usepackage{graphicx}
\graphicspath{ {./images/} }
\usepackage{textcomp}
\usepackage{xcolor}
\usepackage{amssymb}
\usepackage{amsmath}
\usepackage{subcaption}
\usepackage{amsfonts}
\usepackage{amsthm}
\usepackage{gauss}
\usepackage{bm}
\usepackage{chngcntr}
\usepackage{coloremoji}
\usepackage{booktabs} 
\usepackage{tikz}
\usepackage{hyperref}
\usetikzlibrary{shapes,arrows}
  \usetikzlibrary{shapes,shadows}
  \tikzstyle{abstractbox} = [draw=black, fill=white, rectangle,
  inner sep=10pt, style=rounded corners, drop shadow={fill=black,
  opacity=1}]
\tikzstyle{abstracttitle} =[fill=white]
\usetikzlibrary{calc,positioning,shapes.geometric}
\usetikzlibrary{arrows.meta,arrows}

\makeatletter
\pgfdeclareshape{document}{
\inheritsavedanchors[from=rectangle] 
\inheritanchorborder[from=rectangle]
\inheritanchor[from=rectangle]{center}
\inheritanchor[from=rectangle]{north}
\inheritanchor[from=rectangle]{south}
\inheritanchor[from=rectangle]{west}
\inheritanchor[from=rectangle]{east}
\backgroundpath{
\southwest \pgf@xa=\pgf@x \pgf@ya=\pgf@y
\northeast \pgf@xb=\pgf@x \pgf@yb=\pgf@y
\pgf@xc=\pgf@xb \advance\pgf@xc by-10pt 
\pgf@yc=\pgf@yb \advance\pgf@yc by-10pt
\pgfpathmoveto{\pgfpoint{\pgf@xa}{\pgf@ya}}
\pgfpathlineto{\pgfpoint{\pgf@xa}{\pgf@yb}}
\pgfpathlineto{\pgfpoint{\pgf@xc}{\pgf@yb}}
\pgfpathlineto{\pgfpoint{\pgf@xb}{\pgf@yc}}
\pgfpathlineto{\pgfpoint{\pgf@xb}{\pgf@ya}}
\pgfpathclose
\pgfpathmoveto{\pgfpoint{\pgf@xc}{\pgf@yb}}
\pgfpathlineto{\pgfpoint{\pgf@xc}{\pgf@yc}}
\pgfpathlineto{\pgfpoint{\pgf@xb}{\pgf@yc}}
\pgfpathlineto{\pgfpoint{\pgf@xc}{\pgf@yc}}
}
}
\makeatother

\def\BibTeX{{\rm B\kern-.05em{\sc i\kern-.025em b}\kern-.08em
    T\kern-.1667em\lower.7ex\hbox{E}\kern-.125emX}}
\begin{document}

\title{Deep Text Mining of Instagram Data Without Strong Supervision}

\author{
\IEEEauthorblockN{1\textsuperscript{st} Kim Hammar}
\IEEEauthorblockA{
\textit{KTH - Royal Institute of Technology}\\
Stockholm, Sweden \\
kimham@kth.se}
\and
\IEEEauthorblockN{2\textsuperscript{nd} Shatha Jaradat}
\IEEEauthorblockA{
\textit{KTH - Royal Institute of Technology}\\
Stockholm, Sweden \\
shatha@kth.se}
\and
\IEEEauthorblockN{3\textsuperscript{rd} Nima Dokoohaki}
\IEEEauthorblockA{
\textit{KTH - Royal Institute of Technology}\\
Stockholm, Sweden \\
nimad@kth.se}
\and
\IEEEauthorblockN{4\textsuperscript{th} Mihhail Matskin}
\IEEEauthorblockA{
\textit{KTH - Royal Institute of Technology}\\
Stockholm, Sweden \\
misha@kth.se}
}

\maketitle

\begin{abstract}
  With the advent of social media, our online feeds increasingly consist of short, informal, and unstructured text. This textual data can be analyzed for the purpose of improving user recommendations and detecting trends. Instagram is one of the largest social media platforms, containing  both text and images. However, most of the prior research on text processing in social media is focused on analyzing Twitter data, and little attention has been paid to text mining of Instagram data. Moreover, many text mining methods rely on annotated training data, which in practice is both difficult and expensive to obtain. In this paper, we present methods for unsupervised mining of fashion attributes from Instagram text, which can enable a new kind of user recommendation in the fashion domain. In this context, we analyze a corpora of Instagram posts from the fashion domain, introduce a system for extracting fashion attributes from Instagram, and train a deep clothing classifier with weak supervision to classify Instagram posts based on the associated text.

With our experiments, we confirm that word embeddings are a useful asset for information extraction. Experimental results show that information extraction using word embeddings outperforms a baseline that uses Levenshtein distance. The results also show the benefit of combining weak supervision signals using generative models instead of majority voting. Using weak supervision and generative modeling, an $F_1$ score of $0.61$ is achieved on the task of classifying the image contents of Instagram posts based solely on the associated text, which is on level with human performance. Finally, our empirical study provides one of the few available studies on Instagram text and shows that the text is noisy, that the text distribution exhibits the long-tail phenomenon, and that comment sections on Instagram are multi-lingual.
\end{abstract}


\section{Introduction}
Text processing is present in our everyday life and empowers several important utilities, such as, machine translation, web search, personal assistants, and user recommendations. Today, social media is one of the largest sources of text, and while social media fosters the development of a new type of text processing applications, it also brings with it its own set of challenges due to the informal language.

Text in social media is unstructured and has a more informal and conversational tone than text from conventional media outlets \cite{hownoisy}. For instance, text in social media is rich of abbreviations, hashtags, emojis, and misspellings.

Traditional Natural Language Processing (NLP)-tools are designed for formal text and are less effective when applied on informal text from social media \cite{ner_twitter18}. This is why recent research efforts have tried to adapt NLP tools to the social media domain \cite{twitter_pos}. Moreover, methods within the intersection of NLP and machine learning applied to social media have been successful in information extraction \cite{twitter_event}, classification \cite{twitter_class}, and conversation modeling \cite{unsup_twitter}.


Results of the previous work are not enough for our purposes due to the following reasons: (1) many results rely on access to massive quantities of annotated data, something that is not available in our domain; (2) most of the work is focused on Twitter, with little attention to image sharing platforms like Instagram\footnote{\url{Instagram.com}}; and (3) to the best of our knowledge, no prior assessment of complex, multi-label, hierarchical extraction and classification in social media has been made.

Acquisition of annotated data that is accurate and can be used for training text mining models is expensive. Especially in a shifting data domain like social media. In this research, we explore the boundaries of text mining methods that can be effective without this type of strong supervision.

Even if we assume that the main research results from Twitter will be useful in our research on Instagram, we still should take into account several important differences between the two domains. The most prevalent discrepancies are that Instagram is an image-sharing medium while Twitter is a micro-blogging medium, and that Twitter has a character-limit per tweet.

In this paper, we focus on the task of extracting fashion attributes from Instagram posts, and classifying Instagram posts into clothing categories based on the associated text. The work presented in this paper is part of a larger research project. The project aspires to improve the state-of-the-art in fashion recommendation by employing activities in social media and using data crossing multiple domains in the recommendations \cite{shatha_intro}. The text processing methods presented in this paper are meant to be integrated with computer vision models in the project.

Just as other consumption-driven industries, the fashion industry has been influenced by the emergence of social media. Social media is progressively getting more attention by fashion brands and retailers as a source for detecting trends, adapting user recommendations, and for marketing purposes \cite{fashion_sm2}. To give an example, the image-sharing platform Instagram has become a popular medium for fashion branding and community engagement \cite{fashion_article}. This is why extraction and classification of fashion attributes on Instagram is an important task for several modern applications working with user recommendation and detection of fashion trends.

\begin{figure}
\centering
    \scalebox{0.21}{
      \includegraphics{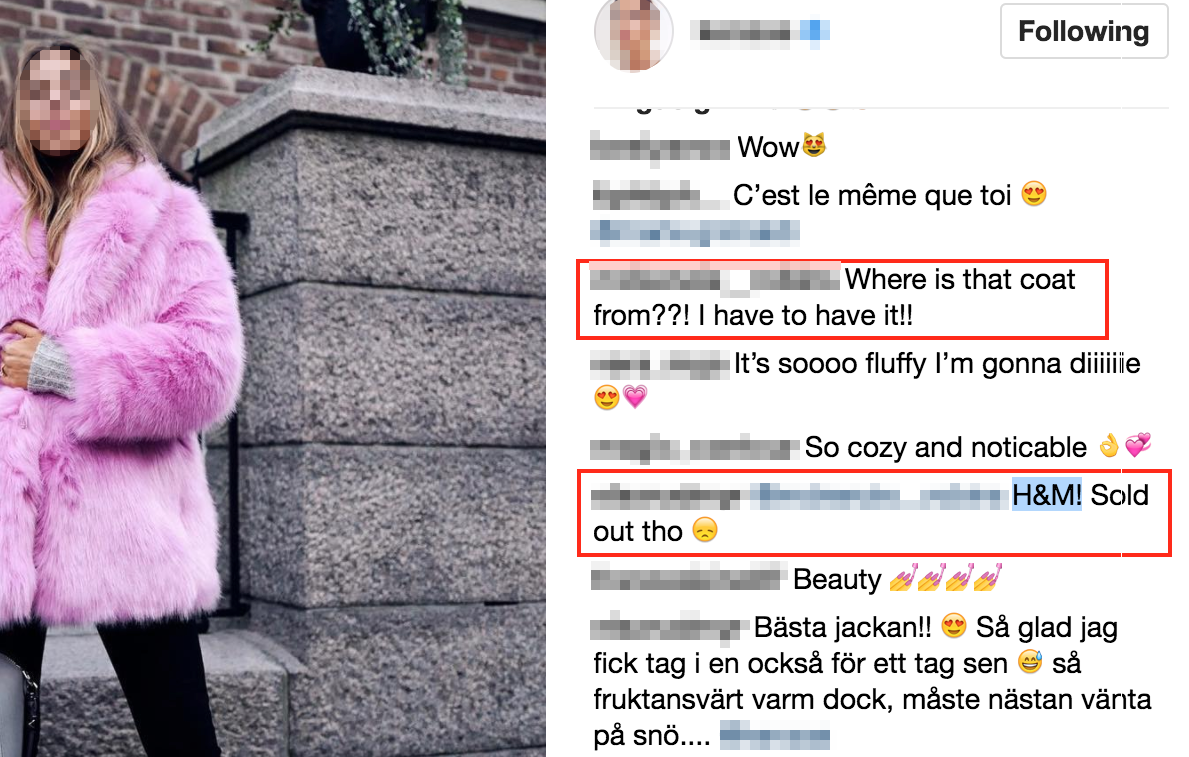}
    }
    \caption{An Instagram post from the fashion community.}
    \label{fig:kenza}
\end{figure}

In addition to hosting images, Instagram contains large volumes of user generated text. Specifically, an Instagram post can be associated with an image caption written by the author of the post, by comments written by other users, and by ``tags'' in the image that refer to other users. Despite being a platform rich of text, little prior work has paid attention to the promising applications of text mining on Instagram. From our case study on Instagram posts in the fashion community, it was revealed that the text often \textit{indicates} the clothing on the associated image, an example of this is given in Fig. \ref{fig:kenza}. We believe that there is a value in the text on Instagram that currently is unutilized. For example, the text on Instagram can be mined and used for predictive modeling and analytics.

Our contribution in this paper includes:
\begin{itemize}
\item An empirical study of Instagram text.
\item A system for unsupervised extraction of fashion attributes from text on Instagram.
\item A novel pipeline for multi-label clothing classification of the text associated with Instagram posts using weak supervision and the \textit{data programming} paradigm \cite{data_prog}.
\end{itemize}
Our empirical study provides one of the few available studies on Instagram text and shows that the text is noisy, that the text distribution exhibits the long-tail phenomenon, and that comment sections on Instagram often are multi-lingual. Moreover, experimental results demonstrate that the use of word embeddings adds semantic word knowledge that is helpful for information extraction and improves the accuracy compared with a baseline that uses Levenshtein distance. Finally, we train a deep text classifier using weak supervision and data programming. The classifier achieves an $F_1$ score of $0.61$ on the task of clothing prediction of Instagram posts based on the text. The accuracy of the classifier is on level with human performance on the task and beats a baseline that uses majority voting.

The rest of this paper is structured as follows. In Section \ref{sec:related_work} we describe related work, and in Section \ref{sec:method} we present our approach to the problem. In Section \ref{sec:setup} we summarize the experimental setup and Section \ref{sec:results} contains the results from our evaluations and our interpretation of the results. Lastly, Section \ref{sec:conc_fw} includes our conclusions and suggestions for future research directions.
\section{Related Work}\label{sec:related_work}
Our research extends prior work on unsupervised information extraction (Section \ref{sec:ie_relwork}) and weakly supervised text classification (Section \ref{sec:ie_weaksup}) working with informal text.
\subsection{Unsupervised Information Extraction}\label{sec:ie_relwork}
In \cite{twitter_event}, the authors propose an approach to event extraction and categorization that uses a supervised tagger to identify events in tweets. Next, the extracted events are categorized using latent variable models, that can make use of unlabeled data. Results demonstrate an improved accuracy compared with a supervised baseline. Their work resembles ours in that they attempt to classify and extract information from noisy text, and try to make use of unlabeled data. However, it has some important differences compared to our setting. In event categorization, the categories are unclear a priori, which fits well into the latent variable model approach. In contrast, our extraction problem has a pre-defined set of classes. Moreover, in their proposed solution, they assume access to an annotated dataset for training a tagger to recognize events in tweets, a corresponding dataset is not available in our domain.

Numerous research efforts have been made on the line of coarse-grained classification in social media using latent variable models \cite{unsup_twitter,ner_twitter18}. These studies differ from our work in two ways. First, most of the work is focused on Twitter. Second, in our research, the goal is a complex multi-label extraction, while the aforementioned work target more general and high-level extraction tasks.

Word embeddings have shown to be a great asset for information extraction. In \cite{ir_we1} the authors evaluate how useful word embeddings are for clinical concept extraction and in \cite{ir_we2} the utility of word embeddings for named entity recognition on Twitter is evaluated. Both results demonstrate improvements when using word embeddings compared to baseline methods.

\subsection{Text Classification with Weak Supervision}\label{sec:ie_weaksup}
For the task of classifying Instagram text, our research builds primarily on results from supervised machine learning. The success of this paradigm of machine learning has traditionally been coupled to annotated datasets. Notable results in supervised text classification are \cite{kim_cnn} and \cite{verydeep_cnn}, both of which differ from our research in that they assume access to a large annotated text corpora for training the classifier.

More recently, weakly supervised approaches have been used for text classification and information extraction. Specifically, the \textit{data programming paradigm} presented in \cite{data_prog}, has achieved promising results. Data programming has been applied to binary and multinomial text extraction and classification tasks \cite{data_prog, snorkel}. To the best of our knowledge, it has neither been applied to multi-label classification tasks, nor to social media text.

\section{Methodology}\label{sec:method}
In text mining, there is a balance between models that rely on domain knowledge and models that rely on annotated training data. In our research, we have experimented with both approaches. In Section \ref{sec:da} we outline how our analysis of the Instagram corpora was performed. Section \ref{sec:ie} describes our second contribution, which is a method for information extraction using an ontology with domain knowledge and word embeddings. Finally, Section \ref{sec:weak_sup} presents the pipeline we used to train a deep text classifier using \textit{weak supervision}. The code for the implementations is publicly available\footnote{\url{https://github.com/shatha2014/FashionRec}}.
\subsection{Empirical Study of Instagram Text}\label{sec:da}
Of special interest in our study was to elucidate how the Instagram text differs from newswire text, as it affects the choice of processing methods. We analyzed a corpora of Instagram posts by measuring the fraction of online-specific tokens, the number of Out-Of-Vocabulary (OOV) words, the number of languages in the corpora, and the text distribution.
\subsection{Extracting Fashion Attributes from Instagram}\label{sec:ie}
\tikzstyle{doc}=[%
draw,
thick,
align=center,
color=black,
shape=document,
minimum width=10mm,
minimum height=10.2mm,
shape=document,
inner sep=2ex,
]
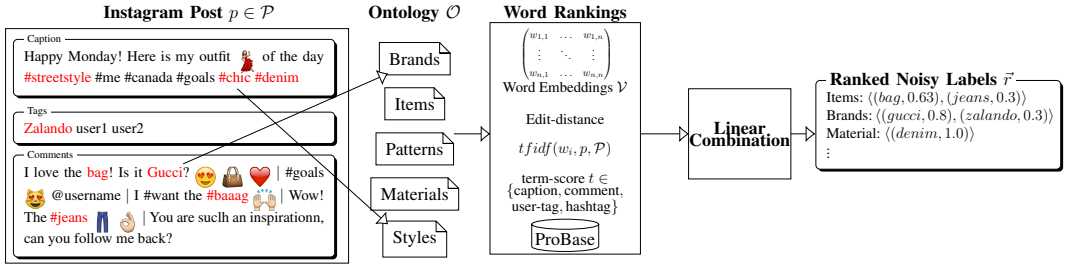
\begin{figure*}[!h]
  \centering
  \scalebox{0.4}{
    \centering
    \input{tikz/ie.tex}
  }
\caption{\textsc{SemCluster}, a system that extracts fashion details from text associated with Instagram posts.}
\label{fig:iepipeline}
\end{figure*}
We have developed a system for extracting fashion attributes from Instagram posts using word embeddings and a domain ontology, subsequently referred to as \textsc{SemCluster}. Fig. \ref{fig:iepipeline} illustrates the workings of the system. The extraction in \textsc{SemCluster} is carried out as follows.
\subsubsection{Text Normalization}\label{semcluster_1}
To begin with, the text of a single post is tokenized with NLTK's \cite{nltk} TweetTokenizer, that is designed to recognize text from social media (a tokenizer that can handle online-specific tokens such as emojis and emoticons). Then the text is normalized by lemmatizing and lower-casing all tokens as well as removing stopwords. Moreover, hashtags, emojis, and user-handles are extracted using regular expressions, and hashtags are segmented using the segmenter presented in \cite{hashtag_segmenter}.
\subsubsection{Ontology Mapping Using Word Embeddings}\label{semcluster_2}
After normalizing the text, it is mapped to a domain ontology that includes fashion brands, items, patterns, materials, and styles. The mapping is based on semantic similarity matching via word embeddings and the cosine similarity metric. Furthermore, each word's contribution to the rankings of the categories in the ontology is scaled by its Term Frequency-Inverse Document Frequency (TF-IDF) score, and its term-score. The term-score has a different weight depending on if the word occurred in the caption, a usertag, a hashtag, or in a comment. After this mapping with the ontology, the $k$ highest ranked entities from the ontology are extracted together with their respective scores.
\subsubsection{Ambiguity Resolution}\label{semcluster_3}
The results are re-ranked based on a source of distant supervision, Probase \cite{probase}. Probase is an API that, for a given word, returns an estimated probability that the word has a certain meaning. For instance, the homonym ``felt'' is both a clothing fabric and a common English word, implying that it will receive a lower rank than a less ambiguous word, like ``polyester''. Hence, Probase is used by \textsc{SemCluster} to resolve word ambiguities.
\subsubsection{Linear Combination}\label{semcluster_4}
The different components in the pipeline are combined in a final ranking $\vec{r}$ through a linear combination defined in \eqref{eq:linear_comb} using the glossary from Table \ref{tab:glossary}.
\begin{table}
\caption{Glossary for \eqref{eq:linear_comb}.}\label{tab:glossary}
\centering
\resizebox{1\columnwidth}{!}{%
\begin{tabular}{llllll} \toprule
  {\textit{Term}} & {\textit{Meaning}} \\ \midrule
  $\mathcal{O}$ & A fashion ontology \\
  $\mathcal{P}$ & Set of all Instagram posts \\
  $p$ & An Instagram post \\
  $cos(\vec{w_i}, \vec{o_j})$ & Cosine similarity between embeddings\\
  $tfidf(w_i,p,\mathcal{P})$ & TF-IDF statistic for word $w_i$ \\
  $h(o_j)$ & Probase lookup of ontology term $o_j$ \\
  $t(w_i)$ & Term-score of word $w_i$ \\
  $\gamma,\eta,\alpha$ & Scaling factors \\
  \bottomrule
\end{tabular}
}
\end{table}
\begin{IEEEeqnarray}{rCl}
\IEEEeqnarraymulticol{3}{l}{
\forall (w_i,o_j)\quad w \in p, o \in \mathcal{O} \quad r(w_i,o_j)}\nonumber\\*\quad
& = & t(w_i) + \gamma h(o_j) + \eta (tfidf(w_i,p,\mathcal{P})) + \alpha (cos(\vec{w_i},\vec{o_j}))
\nonumber\\
 &\vec{r} = & \underset{s_j}{top_k}(\{(o_j, s_j) | o_j \in \mathcal{O} \land s_j = \sum_i r(w_i,o_j)\})\label{eq:linear_comb}
\label{eq:linear_comb}
\end{IEEEeqnarray}

Effectively, the information extraction may be seen as a form of clustering, where clusters are seeded with terms from an ontology, and the $k$ most salient clusters are returned and re-ranked based on distant supervision.
\subsection{Clothing Classification of Instagram Posts}\label{sec:weak_sup}
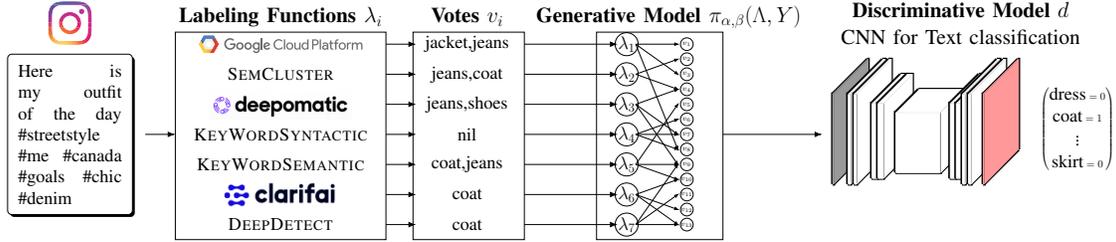
\begin{figure*}[!h]
  \centering
  \scalebox{0.4}{
    \input{tikz/weak_sup.tex}
  }
  \caption[A pipeline for weakly supervised text classification of Instagram posts.]{A pipeline for weakly supervised text classification of Instagram posts\protect.}
\label{fig:weak_sup}
\end{figure*}

This section presents a pipeline for weakly supervised text classification to predict clothing items in Instagram posts. The pipeline is visualized in Fig. \ref{fig:weak_sup} and includes steps devoted to labeling a dataset with weak supervision (Section \ref{sec:labeling}), combining weak labels with data programming to produce probabilistic labels (Section \ref{sec:data_programming}), and training a discriminative model using the probabilistic labels (Section \ref{sec:disc_model}).
\subsubsection{The Classification Task}\label{sec:class_task}
Although multiple classifications are of interest in our research, such as brand classification, and fabric classification, we focus initially on the clothing item classification problem. This task is a multi-label multi-class classification problem with $13$ classes. The classes are as follows: dresses, coats, blouses \& tunics, bags, accessories, skirts, shoes, jumpers \& cardigans, jeans, jackets, tights \& socks, tops \& t-shirts, and trouser \& shorts.
\subsubsection{Data Programming}\label{sec:data_programming}
With the data programming paradigm \cite{data_prog}, weak supervision is encoded with \textit{labeling functions}. A labeling function is any function $\lambda_i: x \rightarrow y$, that takes as input a training example $x$, and outputs a label $y$.  A labeling function is typically realized through some domain heuristic and only labels a subset of the data. Naturally, labels produced by such functions are less accurate than labels produced by human annotators. However, weak labels can be complementary to each other. Several weak labels can be combined with the purpose of obtaining more accurate labels. The innovative part of data programming is the way that it learns a generative model of the labeling process in an unsupervised fashion.

Formally, a labeling function $\lambda_i$ has a probability $\beta$ of labeling an input, and refrain from labeling an input with probability $1-\beta$. Similarly, a labeling function has a probability $\alpha$ of labeling an input correctly. The combination of labeling functions can be modeled as a generative model $\pi_{\alpha,\beta}(\Lambda, Y)$. Where $\Lambda$ is the output matrix after applying all of the labeling functions to the unlabeled data ($\Lambda_{i,j} = \lambda_j(x_i)$), and $Y$ is the true classes, modeled as latent variables.

To train the generative model means to find the parameters of the model that best describe the empirical overlaps that were observed among the labeling functions. In other words, the training process \textit{estimates} the accuracy of each labeling function based on the observed overlaps. Training the generative model can be phrased as an optimization problem that can be solved using maximum likelihood estimation \cite{data_prog}.

Once trained, the parameters learned by the generative model can be used to produce probabilistic training labels $p(Y|\Lambda)$ from the unlabeled data and the output of the labeling functions. When producing the probabilistic labels, more weight is given to accurate labeling functions. Moreover, the uncertainty of each label is encoded by the probability of the label. If labeling functions disagree on a training example, this is encoded as an uncertainty by giving the corresponding label a lower probability. As the labels are probabilistic and not binary, a \textit{noise-aware} loss function is used when training a discriminative model with such labels. A noise-aware loss function is a loss function for minimizing the expected loss with respect to the probabilistic labels.
\subsubsection{Weak Supervision for Fashion Attributes in Instagram Posts}\label{sec:labeling}
We used seven labeling functions to label a dataset of $30$K Instagram posts with fashion attributes. The purpose of using several functions is that we expect that the combination of functions will improve the accuracy of the supervision compared to what each function in isolation would provide. The functions are as follows.
\begin{enumerate}
\item $\lambda_1$, a function that uses Google's Cloud Vision API\footnote{\url{https://cloud.google.com/vision/}} to classify the image associated with the text.
\item $\lambda_2$, the system for information extraction, \textsc{SemCluster}.
\item $\lambda_3$, a function that uses the Deepomatic\footnote{\url{https://www.deepomatic.com/}} API for computer vision to classify the image associated with the text.
\item $\lambda_4$, a function that uses keyword matching to the fashion ontology with Levenshtein distance \cite{leven}.
\item $\lambda_5$, a function that uses keyword matching to the fashion ontology with word embeddings.
\item $\lambda_6$, a function that uses the Clarifai ``Apparel'' model\footnote{\url{https://www.clarifai.com/}} to classify the image associated with the text.
\item $\lambda_7$, a function that uses a pre-trained image-classifier provided by DeepDetect\footnote{\url{https://www.deepdetect.com/}}.
\end{enumerate}
It should be clear that the kind of supervision provided by the aforementioned labeling functions is scalable and extremely cheap in comparison with supervision in the form of human annotations.
\subsubsection{Using Data Programming to Combine Multi-Labels}\label{sec:dp_multi}
In the original data programming paper, a binary classification scenario is studied and it is assumed that labeling functions are binary \cite{data_prog}. To make use of the data programming paradigm for \textit{multi-label} classification, we model the labeling process with one generative model for each class. With this approach, the combination of generative models is able to represent separate accuracy estimates of the labeling functions for each class.

Once the generative models are trained, they are used to produce probabilistic labels $l_{i,j} \in [0,1]$ for each class and training example. The probabilistic labels are produced based on the parameters of the generative models and the outputs of the labeling functions. The probabilistic labels for each class are then combined into a single multi-label by concatenation $\vec{l_i} = \langle l_{i,0}, l_{i,1}, \hdots, l_{i,|C|}\rangle$ (where $C$ denotes the set of classes). After combining the output of the labeling functions into probabilistic multi-labels, they can be used to train a discriminative model in a supervised fashion.
\subsubsection{Discriminative Model}\label{sec:disc_model}
For the discriminative model, we have used the Convolutional Neural Network (CNN) model for text classification presented in \cite{kim_cnn}. This model was chosen as it is established as one of the best performing text classifiers. However, nearly any model could have been used, the only requirement is that the loss function can be modified.

The neural network architecture in \cite{kim_cnn} consists of an embedding input layer, a convolutional layer, and a fully-connected layer of softmax or sigmoid output units. Moreover, the architecture employs max-over-time pooling to detect keywords in the input. The original architecture is designed for the multi-class setting \cite{kim_cnn}. We have extended the network to the multi-label setting working with \textit{probabilistic labels} by switching out the loss function with a noise-aware loss function for multi-label classification. The loss function is defined in \eqref{eq:loss_fun}, where $N$ is the number of classes, $\theta$ is the model parameters, $p(Y_i|\Lambda_i)$ is the probabilistic labels for class $i$, $\sigma$ is the logistic sigmoid function ($\sigma(x) = \frac{1}{1+e^{-x}}$), and $\hat{y_i}$ is the predicted logits for class $i$.
\begin{IEEEeqnarray}{rCl}
\IEEEeqnarraymulticol{3}{l}{
L(\theta) = \frac{1}{N}\sum_{i=0}^N
}\nonumber\\* \quad
&& -(p(Y_i|\Lambda_i)\log(\sigma(\hat{y_i})) + ((1-p(Y_i|\Lambda_i)) \log(1-\sigma(\hat{y_i}))))
\nonumber\\*
\label{eq:loss_fun}
\end{IEEEeqnarray}
\section{Experimental Setup}\label{sec:setup}
This section outlines the experimental setup that was used to produce the results presented in the following section (Section \ref{sec:results}). The experiments include data analysis of an Instagram corpora, evaluating the capability of word embeddings for text mining, and training a deep text classifier.
\subsection{Data}
\subsubsection{Instagram Corpora}
The empirical study of Instagram text was conducted on a provided dataset, consisting of Instagram posts from a community of users in the fashion domain. The data are in the form of a corpora consisting of image captions, user comments, and usertags associated with each post. In entirety, the corpora consists of $143$ accounts, $200$K posts, $9$M comments, and $62$M tokens, out of which $2$M are unique. The numbers were computed before any pre-processing, except applying the NLTK \cite{nltk} TweetTokenizer and removing user-handles.
\subsubsection{Training Dataset}
When training classifiers, a dataset of $30$K Instagram posts annotated with weak labels produced by the labeling functions described in Section \ref{sec:labeling} was used.
\subsubsection{Evaluation Dataset}\label{eval_set}
For evaluation purposes, a smaller annotated dataset of $200$ Instagram posts have been used. The annotation was a collective work by four participants in our research group. Noteworthy is that the truth labels are based on the image associated with the text. In that sense, the evaluation is unfavorable for the text-based analysis. Since the labels are decided by the image, certain posts can have labels that cannot be inferred from the text alone, degrading the measured performance of the developed text mining models.
\subsubsection{Word Embeddings}
The word embeddings used in our experiments were trained on the Instagram corpora. The embeddings were selected after an extrinsic and intrinsic evaluation that included both off-the-shelf embeddings and embeddings trained on the Instagram corpora. The evaluation that compares embeddings is left out in this paper for brevity, readers are advised to \cite{kim_thesis} for details.
\subsection{Data Analysis}
The data analysis was conducted on the entire Instagram corpora. To measure the fraction of emojis, hashtags, and user-handles, the NLTK \cite{nltk} TweetTokenizer was used to tokenize the text, and regular expressions were applied to extract the desirable tokens. To quantify the amount of OOV words, two vocabularies were used, the Google-news vocabulary \cite{googlenews}, and GNU aspell v0.60.7. Finally, \texttt{langid.py} \cite{langid} was used to capture the distribution of languages in the corpora.
\subsection{Unsupervised Information Extraction}
\subsubsection{Evaluation}
The system for information extraction was evaluated by comparing its extraction with the annotated dataset.
\subsubsection{Baseline}
To highlight the utility of word embeddings for information extraction, the built system, \textsc{SemCluster}, was evaluated against a baseline, that we refer to as \textsc{SynCluster}. The baseline follows the same extraction method as \textsc{SemCluster} except that it uses syntactic matching through Levenshtein distance \cite{leven}, instead of the method with word embeddings used in \textsc{SemCluster}.
\subsubsection{Hyperparameters}
In all of the experiments with \textsc{SemCluster}, the term-score was set to $2,1,1,3$ for caption, comments, tags, and hashtags, respectively. With the motivation that we believe that clothing descriptions provided by the author of a post are more accurate than descriptions that occur in user comments. Moreover, the relative weighting among semantic, TF-IDF and Probase was kept equal and $k$ was set to $10$.
\subsubsection{Significance Testing}
When comparing two systems for information extraction, a pairwise t-test on the recorded results was made to measure if the difference between the results is significant. The null-hypothesis in the test was that the results were produced by the same system, and that deviations in the results occurred by chance. The significance testing was done against a p-value threshold of $0.05$.

\subsection{Weakly Supervised Text Classification}
\subsubsection{Evaluation}
Classifiers were evaluated after training by freezing the weights of the models and comparing the models' predictions to the annotated dataset.
\subsubsection{CNN Models and Baselines}
 Four classifiers were evaluated. A deep classification model of the type in \cite{kim_cnn} was trained using the weak labels and the data programming paradigm (\textsc{CNN-DataProgramming}). The same model was also trained with labels obtained by taking the majority vote of the weak labels (\textsc{CNN-MajorityVote}). For training generative models to use in data programming, the Snorkel implementation was used \cite{snorkel}.

A classifier based on the outputs of \textsc{SemCluster} served as a baseline. Moreover, the CNN models were also compared against a human benchmark (\textsc{DomainExpert}). The human benchmark represents the average performance on the classification task of three people from our research group. Human test participants were faced with the same task as the other models, namely to classify Instagram posts based solely on the text.
\subsubsection{Hyperparameters}
Limited hyperparameter tuning was done prior to the experiments. We used $128$ filter windows of size $3$, $4$, and $5$, and a mini-batch size of $256$. Moreover we used a vector dimension in the embedding layer of $300$ with randomly initialized embeddings updated as part of training. For regularization we used a dropout keep probability of $0.7$ and a $l_2$ constraint of $0$. Finally, ReLU ($f(z) = \max(0,z)$) was used as the activation function, and the padding strategy was set to VALID and the learning rate to $0.01$. The values were chosen based on hyperparameter tuning using random-search on $10\%$ randomly chosen examples from the training dataset.

\section{Results and Discussion}\label{sec:results}
In this section, we present our experimental results.
\subsection{What Characterizes Instagram as a Source of Text?}
\subsubsection{Lexical Noise Measurements}
Table \ref{tab:lexical_noise} contains statistics that capture the distinctive properties of the Instagram corpora compared with newswire text. Removing all online-specific tokens (hashtags, user-handles, emojis, URLs) results in an OOV fraction of $0.30$ based on the aspell dictionary, that can be compared with $0.25$ that was obtained by \cite{hownoisy} on a Twitter corpora using the same pre-processing and dictionary.

\begin{table}
\caption{Measurements of lexical noise in the corpora.}\label{tab:lexical_noise}
\centering
\resizebox{1\columnwidth}{!}{%
\begin{tabular}{llllll} \toprule
    {\textit{Text Statistic}} & {\textit{Fraction of corpora size}} & {\textit{Average/post}}\\ \midrule
  Emojis  & $0.15$ & $48.63$  \\
  Hashtags  & $0.03$ & $9.14$  \\
  User-handles  & $0.06$ & $18.62$ \\
  Google-OOV words  & $0.46$ & $145.02$   \\
  Aspell-OOV words  & $0.47$ & $147.61$   \\
  \bottomrule
\end{tabular}
}
\end{table}
\subsubsection{Language Distribution}
Although all Instagram posts in the corpora are from English accounts, the comments sections are often multi-lingual. Applying \texttt{langid.py} \cite{langid} on the set of $9$ million comments reveals that $52$\% of the comments are primarily written in English. The Language identified as the second most common was Chinese on $6.5$\%, followed by Japanese on $5$\%, German on $3$\%, and Spanish on $2$\%. In total, $97$ languages were identified in the set of comments.
\subsubsection{Text Distributions}
\begin{figure}
  \centering
    \scalebox{0.55}{
      \includegraphics{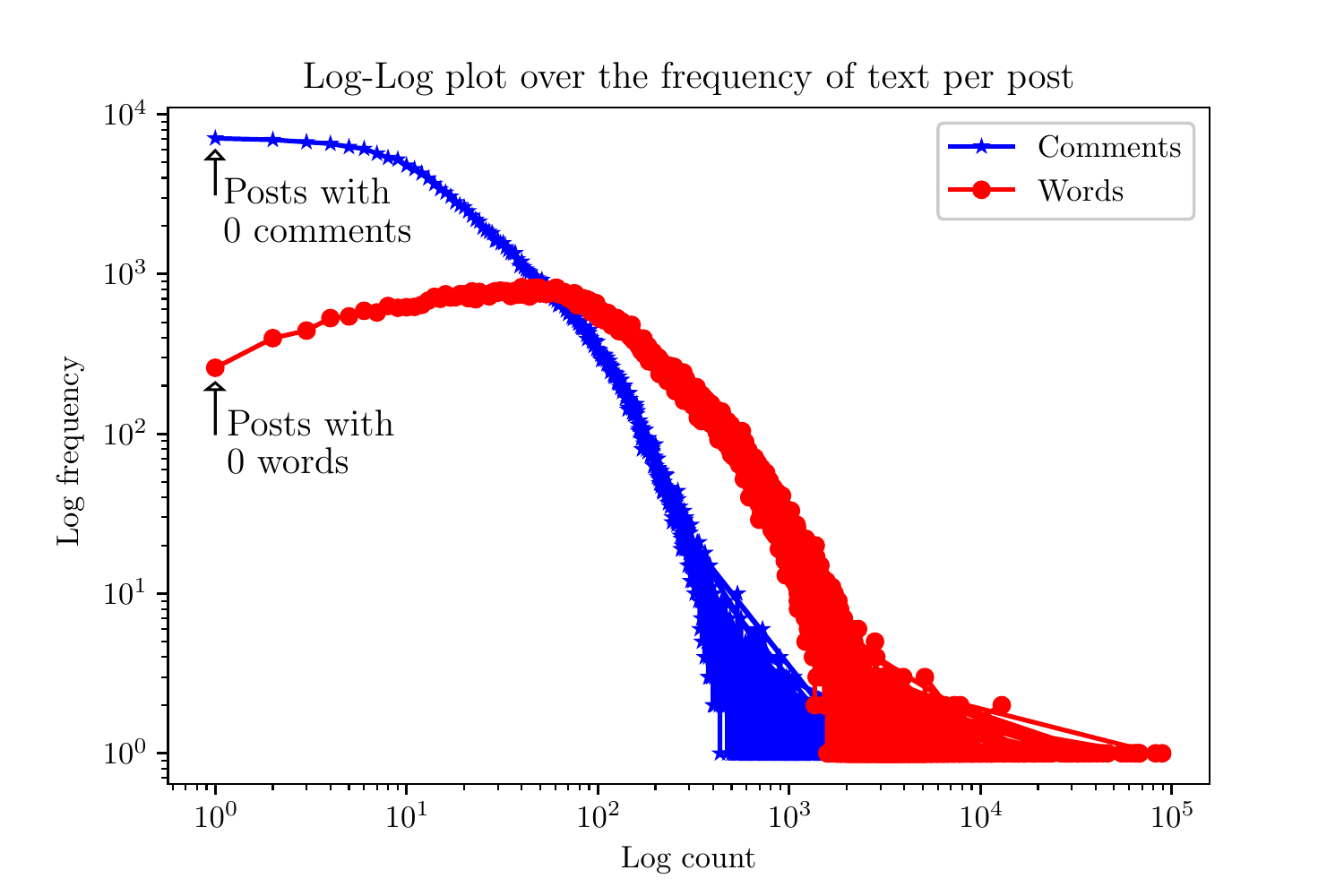}
    }
    \caption{The text distribution in the corpora.}
    \label{fig:text_distributions}
\end{figure}

The number of comments associated with Instagram posts is varying. Data analysis indicate that the distribution of comments and amount of text associated with posts exhibit the long tail phenomenon, and the frequencies of number of comments roughly follows a \textit{power law} relationship (Fig. \ref{fig:text_distributions}). Some posts have no comments at all, while other posts have a few thousand comments. The mean length of captions and comments in the corpora is $29$, and $6$ tokens, respectively.

\subsubsection{Discussion}
In comparison with measurements on Twitter corpora \cite{hownoisy}, text from Instagram is just as noisy based on our measurements (Table \ref{tab:lexical_noise}). Notable is also the high diversity of languages occurring in the comment sections on Instagram and the short length of comments (mean length measured to be 6 tokens).

The long-tail distribution of text on Instagram can be explained with the follower count of the post author and the preferential attachment theory \cite{pfa}. As an Instagram post attracts a lot of comments, it will get a larger spread on the Instagram platform. This causes a snowball effect, where a post that already has many comments will be more likely to attract even more comments.
\subsection{How Effective Are Word Embeddings for Information Extraction in Social Media?}
\subsubsection{Comparison with Baseline}
A comparison between \textsc{SemCluster} and the baseline, \textsc{SynCluster}, is presented in Table \ref{tab:ie_eval}. \textsc{SemCluster} beats the baseline in four out of the five sub-tasks.
\begin{table*}
\caption{Performance comparison between \textsc{SemCluster} and \textsc{SynCluster}. Significant performance degradation of the baseline, \textsc{SynCluster}, in comparison to \textsc{SemCluster} is denoted with ($-$), with p-value $\leq 0.05$.}\label{tab:ie_eval}
\centering
\resizebox{1.45\columnwidth}{!}{%
\begin{tabular}{llllllllll} \toprule
   {\textit{Method/Category}} & {\textit{NDGC@1}} & {\textit{NDGC@3}} & {\textit{NDGC@5}} & {\textit{NDGC@10}} & {\textit{P@1}} & {\textit{P@3}} & {\textit{P@5}} & {\textit{P@10}} & {\textit{MAP}} \\ \midrule
  \textsc{SemCluster}/Item  & $\bm{0.833}$ & $\bm{0.658}$ & $\bm{0.691}$ & $\bm{0.807}$ & $\bm{0.833}$ & $\bm{0.546}$ & $\bm{0.454}$ & $\bm{0.309}$ & $\bm{0.733}$ \\
  \textsc{SynCluster}/Item  & $0.781$ & $0.581^-$ & $0.607^-$ & $0.767^-$ & $0.781$ & $0.474^-$ & $0.370^-$ & $0.296$ & $0.641^-$ \\
  \midrule
  \textsc{SemCluster}/Style  & $\bm{0.399}$ & $\bm{0.505}$ & $\bm{0.519}$ & $\bm{0.548}$ & $\bm{0.417}$ & $\bm{0.204}$ & $\bm{0.139}$ & $\bm{0.069}$ & $\bm{0.539}$ \\
  \textsc{SynCluster}/Style  & $0.367$ & $0.415^-$ & $0.425^-$ & $0.507$ & $0.367$ & $0.130^-$ & $0.123$ & $0.069$ & $0.474^-$ \\
  \midrule
  \textsc{SemCluster}/Pattern  & $0.087$ & $0.179$ & $0.353$ & $0.444$ & $0.087$ & $0.110$ & $0.169$ & $\bm{0.118}$ & $0.296$ \\
  \textsc{SynCluster}/Pattern  & $\bm{0.108}$ & $\bm{0.413}$ & $\bm{0.498}$ & $\bm{0.512}$ & $\bm{0.108}$ & $\bm{0.221}$ & $\bm{0.193}$ & $\bm{0.117}$ & $\bm{0.395}$ \\
  \midrule
  \textsc{SemCluster}/Material  &$\bm{0.296}$ & $\bm{0.286}$ & $\bm{0.324}$ & $\bm{0.393}$ & $\bm{0.286}$ & $\bm{0.264}$ & $\bm{0.233}$ & $\bm{0.165}$ & $\bm{0.373}$ \\
  \textsc{SynCluster}/Material  & $0.113^-$ & $0.104^-$ & $0.137^-$ & $0.209^-$ & $0.113^-$ & $0.107^-$ & $0.109^-$ & $0.092^-$ & $0.227^-$ \\
  \midrule
  \textsc{SemCluster}/Brand  & $\bm{0.062}$ & $\bm{0.066}$ & $\bm{0.062}$ & $\bm{0.064}$ & $\bm{0.032}$ & $\bm{0.056}$ & $\bm{0.036}$ & $\bm{0.039}$ & $\bm{0.194}$ \\
  \textsc{SynCluster}/Brand  & $0.016$ & $0.010$ & $0.010$ & $0.010$ & $0.016$ & $0.005$ & $0.003$ & $0.002$ & $0.159$ \\
  \bottomrule
\end{tabular}
}
\end{table*}
\subsubsection{Error Analysis}\label{ie_error}
The main cause of error in the extraction is text sparsity. Since the system relies solely on text for information extraction, its performance degrades when the text is insufficient. The aforementioned problem is the main reason that extracting brands is harder than extracting clothing items, as brands are rarely mentioned in the text. Additionally, before introducing Probase for word disambiguation, extraction of homonym words was an issue.

The baseline, \textsc{SynCluster}, performs comparable with \textsc{SemCluster} on posts that contain words that have direct mappings to words in the fashion ontology. However, for posts where the clothing details is not as obvious to infer from the text, the performance of \textsc{SynCluster} degrades in comparison with \textsc{SemCluster}.


\subsubsection{Discussion}
We believe that word embeddings are particularly useful in the social media domain where there is a high syntactic variety in the text (many languages and different spellings). The high syntactic variety makes it difficult to use syntactic word similarity for information extraction. This is manifested in the results of Table \ref{tab:ie_eval}, where \textsc{SemCluster} outperformed the baseline, \textsc{SynCluster}. The intuition behind this result is that word embeddings can semantically relate the Instagram text to the ontology. This enables the system to also extract information that is not explicitly mentioned in the ontology.

\subsection{How Useful Is Weak Supervision for Training a Classifier?}
\subsubsection{The Data Programming Paradigm Versus Majority Voting}
Table \ref{tab:weak_sup} compares results from the CNN model trained with weak labels combined through majority voting with results from the same model trained with probabilistic labels obtained with data programming. The data programming approach achieves the best $F_1$ result, on level with the human benchmark, beating both \textsc{SemCluster} and \textsc{CNN-MajorityVote}. The human benchmark had a higher precision but a lower recall than the CNN models.
\begin{table}
\caption{The average performance from three training runs.}\label{tab:weak_sup}
\centering
\resizebox{\columnwidth}{!}{%
\begin{tabular}{llllllllll} \toprule
  {\textit{Model}} & {\textit{Accuracy}} & {\textit{Precision}} & {\textit{Recall}} & {\textit{$F_1$}} \\ \midrule
  \textsc{CNN-DataProgramming} & $0.797 \pm 0.01$ & $0.566 \pm 0.05$ & $0.678 \pm 0.04$ & $\bm{0.616} \pm 0.02$\\
 \textsc{CNN-MajorityVote} & $0.739 \pm 0.02$ & $0.470 \pm 0.06$ & $\bm{0.686} \pm 0.05$ & $0.555 \pm 0.03$\\
  \textsc{SemCluster} & $0.719$ & $0.541$ & $0.453$ & $0.493$ \\
  \textsc{DomainExpert} & $\bm{0.807}$ & $\bm{0.704}$ & $0.529$ & $0.604$\\
  \bottomrule
\end{tabular}
}
\end{table}
\subsubsection{Generative Models of the Labeling Functions}
\begin{figure}
  \centering
    \scalebox{0.45}{
      \includegraphics{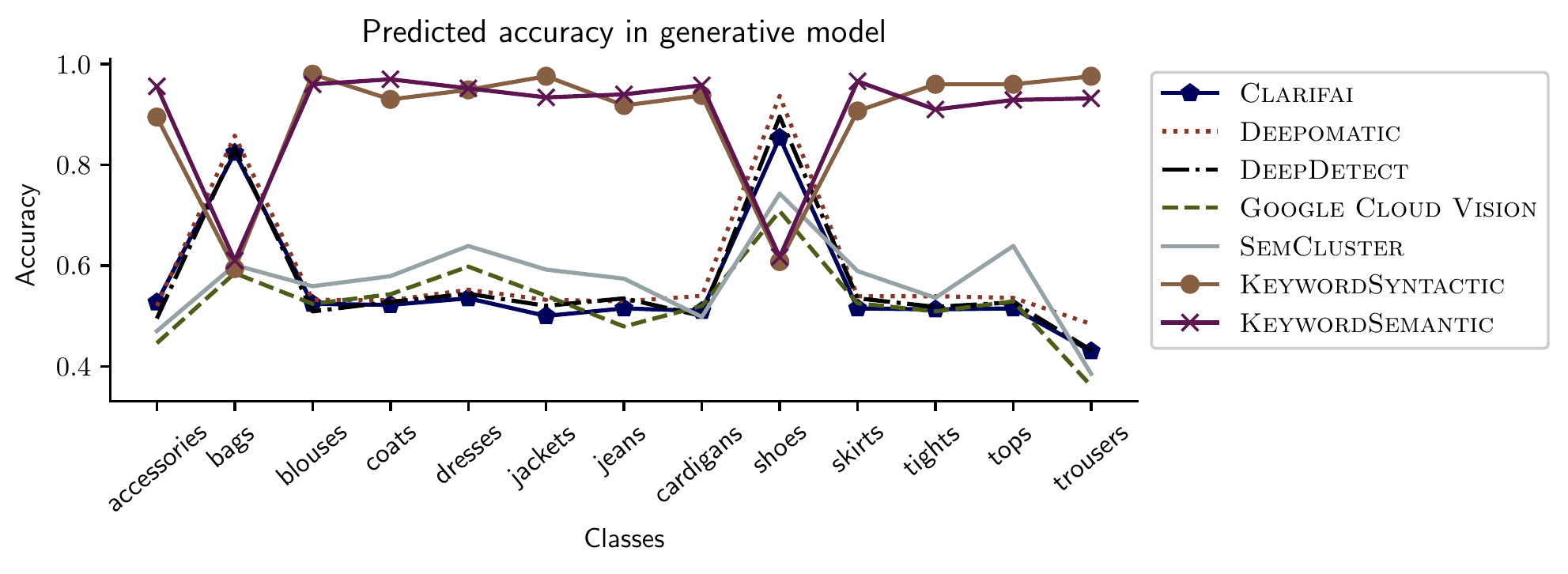}
    }
    \caption{Accuracy of labeling functions in generative models.}
    \label{fig:lf_accs}
\end{figure}

\sloppy Fig. \ref{fig:lf_accs} visualizes the relative accuracy between labeling functions that was learned by the generative models in \textsc{CNN-DataProgramming}. The keyword-functions were given the highest accuracy overall, indicating that when the keywords are found in the text it tend to be telling for the image contents. This implies that the keyword functions often agrees with the majority in their votes, which in turn gives them a high estimated accuracy. In general, the relative accuracy among labeling functions differed from class to class. The spikes in the accuracy of \textsc{Clarifai}, \textsc{Deepomatic}, and \textsc{DeepDetect} on the classes of ``bags'' and ``shoes'' indicate  that the APIs are especially consistent in their predictions on those classes.

\subsubsection{Error Analysis}
As discussed in Section \ref{ie_error}, a part of the error is attributable to the disparity between the labels in the test set, and the text. As the ground truth is determined based on the image contents, there is an inherent error when information is lacking in the text. This is also evident from the relatively low human benchmark on the task ($0.60$ $F_1$).

\subsubsection{Discussion}
Considering that not all clothing items can be inferred from the text and that the human benchmark on the task is $0.60$, the achieved $F_1$ score of $0.61$ is promising. A substantial improvement is to be expected when integrating the text classifier with a model analyzing the image contents.

Combining multiple signals of weak supervision improves the accuracy compared to the baseline system for information extraction. Additionally, when combining the labels by using generative models, rather than majority voting, an increase of six $F_1$ points was observed. The results are concordant with prior work using data programming \cite{data_prog,snorkel}. This result indicates that when combining labels using majority voting, potential signals to learn from is lost.

In \cite{data_prog} it is assumed that labeling functions are binary. We propose to extend the base model to the multi-label scenario by learning a separate generative model for each class. In our experiments, the relative accuracy of labeling functions differed between classes, strengthening our belief that learning separate generative models for each class is useful.

\section{Conclusion and Future Work}\label{sec:conc_fw}
In this paper we presented the first empirical study of Instagram text that we are aware of. Moreover, we presented two systems for text mining of Instagram text without access to strong supervision. The results demonstrate that the text on Instagram is just as noisy as have been reported in studies on Twitter text, that the text distribution has a long tail, and that the comment sections on Instagram are multi-lingual. We also confirmed the capabilities of word embeddings for information extraction and that weak supervision is a viable approach for training deep models with unlabeled data. With weak supervision, we were able to label a large dataset in hours, something that would have taken months to do with human annotators.

In future work we plan to combine the text mining methods presented in this paper with a model that analyzes the image contents associated with the text.
\bibliographystyle{IEEEtran}
\bibliography{IEEEabrv,references.bib}

\end{document}

%% file: tikz/ie.tex
      \begin{tikzpicture}[fill=white, >=stealth,
    node distance=0cm,
    database/.style={
      cylinder,
      cylinder uses custom fill,
      shape border rotate=90,
      aspect=0.15,
      draw}]
        \node [abstractbox] at (-3.5,5.7) (caption)
        {\begin{minipage}{0.55\linewidth}
            \Large Happy Monday! Here is my outfit \coloremoji{💃} of the day \textcolor{red}{\#streetstyle} \#me \#canada \#goals \textcolor{red}{\#chic \#denim}
          \end{minipage}};
        \node[abstracttitle, right=10pt] at (caption.north west) {Caption};

\node[abstractbox] at (-3.5,3.7) (tags)
        {\begin{minipage}{0.55\linewidth}
           \Large \textcolor{red}{Zalando} user1 user2
          \end{minipage}};
        \node[abstracttitle, right=10pt] at (tags.north west) {Tags};

                \node[abstractbox] at (-3.5,1.1) (comments)
        {\begin{minipage}{0.55\linewidth}
            \Large I love the \textcolor{red}{bag}! Is it \textcolor{red}{Gucci}? \coloremoji{😍} \coloremoji{👜} \coloremoji{❤} $|$ \#goals \coloremoji{😻} @username $|$ I \#want the \textcolor{red}{\#baaag} \coloremoji{🙌} $|$ Wow! The \textcolor{red}{\#jeans} \coloremoji{👖} \coloremoji{👌} $|$ You are suclh an inspirationn, can you follow me back?
          \end{minipage}};
        \node[abstracttitle, right=10pt] at (comments.north west) {Comments};

        \node at (4.5, 7.5)   (a) {\LARGE \textbf{Ontology} $\mathcal{O}$};
        \node[doc] at (4.5, 6) (x) {\LARGE Brands};
        \node[doc] at (4.5, 4.5) (x) {\LARGE Items};
        \node[doc] at (4.5, 3) (x) {\LARGE Patterns};
        \node[doc] at (4.5, 1.5) (x) {\LARGE Materials};
        \node[doc] at (4.5, 0) (x) {\LARGE Styles};

        \draw[-] (-9.1,7) rectangle (2.3,-0.8);
        \node at (-3, 7.5)   (igpost) {\LARGE\textbf{Instagram Post} $p \in \mathcal{P}$};

        \draw [-{Latex[length=3mm,width=5mm]}] (-3.2,2.5) -- (3.6,5.5);
        \draw [-{Latex[length=3mm,width=5mm]}] (-1.4,5.0) -- (3.6,0.5);
        \draw [-{Latex[length=3mm,width=5mm]}] (5.8,3.5) -- (7,3.5);
        \node[database] (probase) at (9.5,0) {\LARGE ProBase};

        \node at (9.5, 7.5)   (a) {\LARGE \textbf{Word Rankings}};
        \node [text width=3cm] at (9.5,6)
{
  \begin{align*}
  \begin{gmatrix}[p]
    w_{1,1} & \hdots  & w_{1,n}\\
    \vdots & \ddots & \vdots \\
    w_{n,1} & \hdots & w_{n,n}
  \end{gmatrix}\\
\end{align*}
};
\node at (9.5, 5)   (vecs) {\Large Word Embeddings $\mathcal{V}$};
\node at (9.5, 4)   (lev) {\Large Edit-distance};
\node at (9.5, 3)   (freq) {\Large $tfidf(w_i,p,\mathcal{P})$};
\node[align=center] at (9.5, 1.5)   (termscore) {\Large term-score $t \in$\\\Large $\{\text{caption}, \text{comment},$ \\\Large$\text{user-tag}, \text{hashtag}\}$};

\draw[-] (7,-0.5) rectangle (12,7.2);
\draw [-{Latex[length=3mm,width=5mm]}] (12,3.5) -- (13.6,3.5);
\node[align=center] at (15.3, 3.5)   (combine) {\LARGE\textbf{Linear}\\ \LARGE\textbf{Combination}};
\draw[-] (13.6,2) rectangle (17,5);

\draw [-{Latex[length=3mm,width=5mm]}] (17,3.5) -- (17.8,3.5);
\node[abstractbox] at (21.9,3.8) (results)
        {\begin{minipage}{0.41\linewidth}
            \Large Items: $\langle (bag, 0.63), (jeans, 0.3)\rangle$\\
            \Large Brands: $\langle (gucci, 0.8), (zalando, 0.3) \rangle$\\
            \Large Material: $\langle (denim, 1.0) \rangle$\\
            $\vdots$
          \end{minipage}};
        \node[abstracttitle, right=10pt] at (results.north west) {\LARGE\textbf{Ranked Noisy Labels} $\vec{r}$};
\end{tikzpicture}

%% file: tikz/weak_sup.tex
\tikzset{%
  block/.style    = {draw, thick, rectangle, minimum height = 3em,
    minimum width = 3em},
  sum/.style      = {draw, circle, node distance = 2cm}, 
  input/.style    = {coordinate}, 
  output/.style   = {coordinate} 
}
\usetikzlibrary{shapes,shadows}
\usetikzlibrary{arrows.meta}
  \tikzstyle{abstractbox} = [draw=black, fill=white, rectangle,
  inner sep=10pt, style=rounded corners, drop shadow={fill=black,
  opacity=1}]
\tikzstyle{abstracttitle} =[fill=white]
\begin{tikzpicture}[auto, thick, node distance=2cm, >=triangle 45]
\node[inner sep=0pt] (ig) at (-6.5,1.7)
{\includegraphics[width=.08\textwidth]{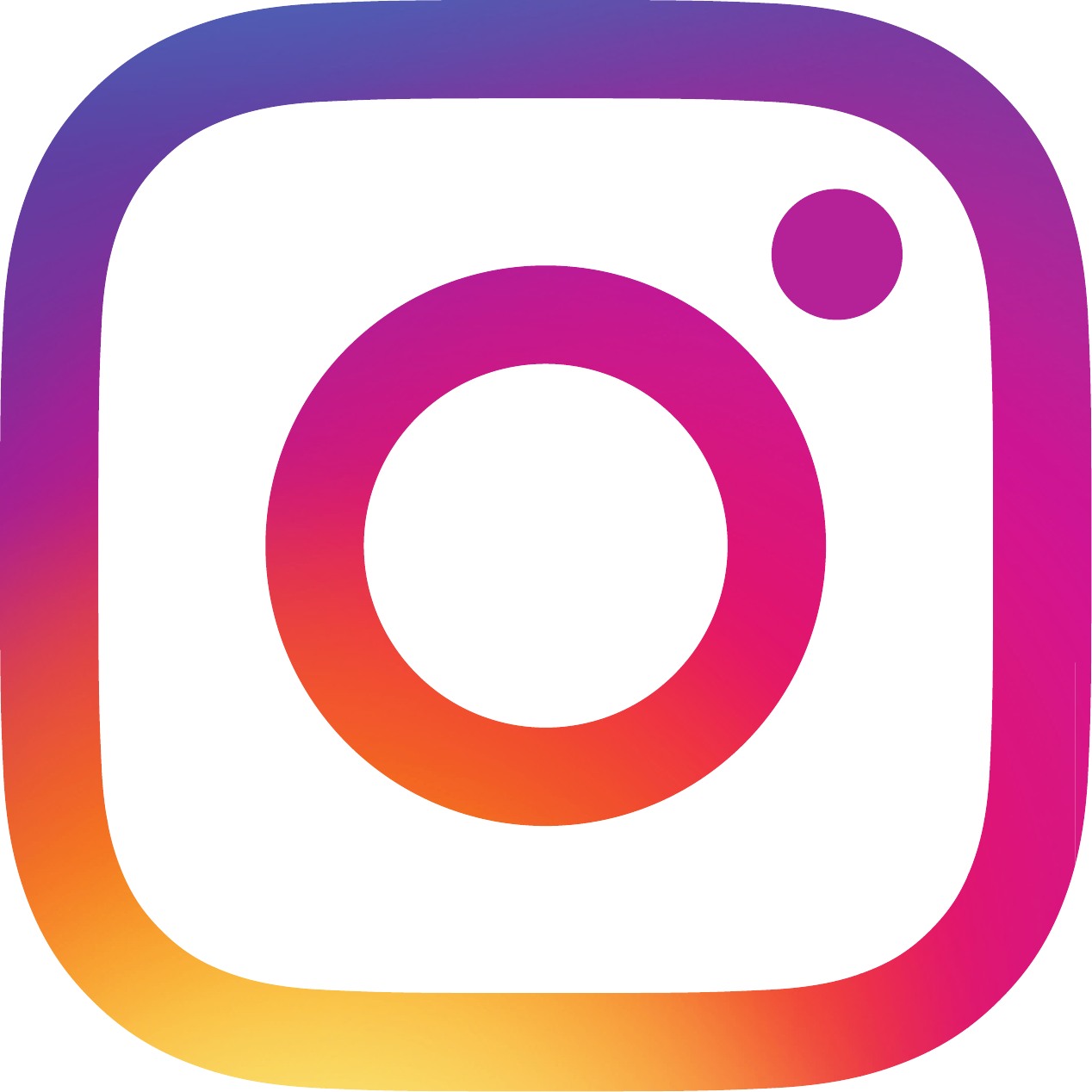}};

\node [abstractbox] at (-6.5,-2) (caption)
{\begin{minipage}{0.19\linewidth}
\LARGE Here is my outfit of the day \#streetstyle \#me \#canada \#goals \#chic \#denim
\end{minipage}};

\draw [-{Latex[length=2.5mm]}] (-4,-2) -- (-3,-2);

\node[inner sep=0pt] (lftitle) at (0.5,1.9)
{\huge\textbf{Labeling Functions $\lambda_i$}};
\node[inner sep=0pt] (google) at (0.5,1)
{\includegraphics[width=.3\textwidth]{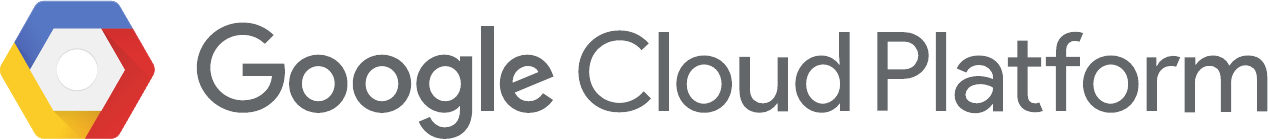}};
\node[inner sep=0pt] (semcluster) at (0.5,0)
{\LARGE\textsc{SemCluster}};
\node[inner sep=0pt] (deepomatic) at (0.5,-1)
{\includegraphics[width=.26\textwidth]{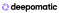}};
\node[inner sep=0pt] (keywordsyntacticmatch) at (0.5,-2)
{\LARGE\textsc{KeyWordSyntactic}};
\node[inner sep=0pt] (clarifai) at (0.5,-4)
{\includegraphics[width=.22\textwidth]{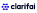}};
\node[inner sep=0pt] (keywordsemanticmatch) at (0.5,-3)
{\LARGE\textsc{KeyWordSemantic}};
\node[inner sep=0pt] (deepdetect) at (0.5,-5)
{\LARGE\textsc{DeepDetect}};

\draw[-] (-3,1.4) rectangle (4.0,-5.5);



		\newcommand{\networkLayer}[6]{
			\def\a{#1} 
			\def\b{0.02}
			\def\c{#2} 
			\def\t{#3} 
			\ifthenelse {\equal{#6} {}} {\def\y{0}} {\def\y{#6}} 

			\draw[line width=1mm](\c+\t,\y,0) -- (\c+\t,\y+\a,0) -- (\t,\y+\a,0);                                                      
			\draw[line width=1mm](\t,\y,\a) -- (\c+\t,\y,\a) node[midway,below] {#5} -- (\c+\t,\y+\a,\a) -- (\t,\y+\a,\a) -- (\t,\y,\a); 
			\draw[line width=1mm](\c+\t,\y,0) -- (\c+\t,\y,\a);
			\draw[line width=1mm](\c+\t,\y+\a,0) -- (\c+\t,\y+\a,\a);
			\draw[line width=1mm](\t,\y+\a,0) -- (\t,\y+\a,\a);
			\filldraw[#4] (\t+\b,\y,\a) -- (\c+\t-\b,\y,\a) -- (\c+\t-\b,\y+\a-\b,\a) -- (\t+\b,\y+\a-\b,\a) -- (\t+\b,\y+\b,\a); 
			\filldraw[#4] (\t+\b,\y+\a,\a-\b) -- (\c+\t-\b,\y+\a,\a-\b) -- (\c+\t-\b,\y+\a,\b) -- (\t+\b,\y+\a,\b);

			\ifthenelse {\equal{#4} {}}
			{} 
			{\filldraw[#4] (\c+\t,\y,\a-\b) -- (\c+\t,\y+\b,\b) -- (\c+\t,\y+\a-\b,\b) -- (\c+\t,\y+\a-\b,\a-\b);} 
		}

		\networkLayer{3.0}{0.03}{20}{color=gray!80}{}{-2.5}

		\networkLayer{3.0}{0.1}{20.5}{color=white}{}{-2.5}    
		\networkLayer{3.0}{0.1}{20.7}{color=white}{}{-2.5}        
		\networkLayer{2.5}{0.2}{21.1}{color=white}{}{-2.5}    
		\networkLayer{2.5}{0.2}{21.3}{color=white}{}{-2.5}        
		\networkLayer{2.0}{1.5}{21.7}{color=white}{}{-2.5}    

		\networkLayer{2.5}{0.2}{23.7}{color=white}{}{-2.5}        
		\networkLayer{2.5}{0.2}{23.9}{color=white}{}{-2.5}        
		\networkLayer{3.0}{0.1}{24.3}{color=white}{}{-2.5}        
		\networkLayer{3.0}{0.1}{24.5}{color=white}{}{-2.5}        

		\networkLayer{3.0}{0.03}{25}{color=red!40}{}{-2.5}          
\node[text width=3cm] (output image) at (27.1,-2.1) {
  \begin{align*}
  \begin{gmatrix}[p]
    \LARGE\text{dress}=0 \\
    \LARGE\text{coat}=1 \\
    \LARGE\vdots \\
    \LARGE\text{skirt}=0
  \end{gmatrix}\\
\end{align*}
};
\node[inner sep=0pt] (votestitle) at (6.7,1.9)
{\huge\textbf{Votes $v_i$}};
\node[inner sep=0pt] (googlevote) at (6.7,1)
{\LARGE jacket,jeans};
\draw [-{Latex[length=2.5mm]}] (4,1) -- (5,1);
\node[inner sep=0pt] (semclustervote) at (6.7,0)
{\LARGE jeans,coat};
\draw [-{Latex[length=2.5mm]}] (4,0) -- (5,0);
\node[inner sep=0pt] (deepomaticvote) at (6.7,-1)
{\LARGE jeans,shoes};
\draw [-{Latex[length=2.5mm]}] (4,-1) -- (5,-1);
\node[inner sep=0pt] (keywordsyntacticmatchvote) at (6.7,-2)
{\LARGE nil};
\draw [-{Latex[length=2.5mm]}] (4,-2) -- (5,-2);
\node[inner sep=0pt] (clarifaivote) at (6.7,-3)
{\LARGE coat,jeans};
\draw [-{Latex[length=2.5mm]}] (4,-3) -- (5,-3);
\node[inner sep=0pt] (keywordsemanticmatchvote) at (6.7,-4)
{\LARGE coat};
\draw [-{Latex[length=2.5mm]}] (4,-4) -- (5,-4);
\node[inner sep=0pt] (deepdetectvote) at (6.7,-5)
{\LARGE coat};
\draw [-{Latex[length=2.5mm]}] (4,-5) -- (5,-5);
\draw[-] (4.9,1.4) rectangle (8.6,-5.5);

\draw [-{Latex[length=2.5mm]}] (8.6,1) -- (11.7,1);
\draw [-{Latex[length=2.5mm]}] (8.6,0) -- (11.7,0);
\draw [-{Latex[length=2.5mm]}] (8.6,-1) -- (11.7,-1);
\draw [-{Latex[length=2.5mm]}] (8.6,-2) -- (11.7,-2);
\draw [-{Latex[length=2.5mm]}] (8.6,-3) -- (11.7,-3);
\draw [-{Latex[length=2.5mm]}] (8.6,-4) -- (11.7,-4);
\draw [-{Latex[length=2.5mm]}] (8.6,-5) -- (11.7,-5);

\node[inner sep=0pt] (votestitle) at (13.4,1.9)
{\huge\textbf{Generative Model $\pi_{\alpha, \beta}(\Lambda, Y)$}};

\node[circle,draw,inner sep=0pt] (a) at (12,1) {\LARGE$\lambda_1$};
\node[circle,draw,inner sep=0pt] (a) at (12,0) {\LARGE$\lambda_2$};
\node[circle,draw,inner sep=0pt] (a) at (12,-1) {\LARGE$\lambda_3$};
\node[circle,draw,inner sep=0pt] (a) at (12,-2) {\LARGE$\lambda_4$};
\node[circle,draw,inner sep=0pt] (a) at (12,-3) {\LARGE$\lambda_5$};
\node[circle,draw,inner sep=0pt] (a) at (12,-4) {\LARGE$\lambda_6$};
\node[circle,draw,inner sep=0pt] (a) at (12,-5) {\LARGE$\lambda_7$};

\draw [-{Latex[length=1mm]},line width=0.001mm] (12.3,1) -- (13.7,1);
\draw [-{Latex[length=1mm]},line width=0.001mm] (12.3,1) -- (13.7,0);
\draw [-{Latex[length=1mm]},line width=0.001mm] (12.3,1) -- (13.7,-2.5);
\draw [-{Latex[length=1mm]},line width=0.001mm] (12.3,0) -- (13.7,0.5);
\draw [-{Latex[length=1mm]},line width=0.001mm] (12.3,0) -- (13.7,-0.5);
\draw [-{Latex[length=1mm]},line width=0.001mm] (12.3,-1) -- (13.7,-3);
\draw [-{Latex[length=1mm]},line width=0.001mm] (12.3,-1) -- (13.7,-2);
\draw [-{Latex[length=1mm]},line width=0.001mm] (12.3,-1) -- (13.7,-0.5);
\draw [-{Latex[length=1mm]},line width=0.001mm] (12.3,-2) -- (13.7,-2);
\draw [-{Latex[length=1mm]},line width=0.001mm] (12.3,-2) -- (13.7,-2.5);
\draw [-{Latex[length=1mm]},line width=0.001mm] (12.3,-2) -- (13.7,-1.5);
\draw [-{Latex[length=1mm]},line width=0.001mm] (12.3,-3) -- (13.7,-1);
\draw [-{Latex[length=1mm]},line width=0.001mm] (12.3,-3) -- (13.7,-3);
\draw [-{Latex[length=1mm]},line width=0.001mm] (12.3,-3) -- (13.7,-3.5);
\draw [-{Latex[length=1mm]},line width=0.001mm] (12.3,-4) -- (13.7,-5);
\draw [-{Latex[length=1mm]},line width=0.001mm] (12.3,-4) -- (13.7,-4.5);
\draw [-{Latex[length=1mm]},line width=0.001mm] (12.3,-4) -- (13.7,-3.5);
\draw [-{Latex[length=1mm]},line width=0.001mm] (12.3,-5) -- (13.7,-5);
\draw [-{Latex[length=1mm]},line width=0.001mm] (12.3,-5) -- (13.7,-4);
\draw [-{Latex[length=1mm]},line width=0.001mm] (12.3,-5) -- (13.7,-3);

\node[circle,draw,inner sep=0pt,minimum size=12pt] (a) at (14,1) {\scriptsize$v_1$};
\node[circle,draw,inner sep=0pt,minimum size=12pt] (a) at (14,0.5) {\scriptsize$v_2$};
\node[circle,draw,inner sep=0pt,minimum size=12pt] (a) at (14,0) {\scriptsize$v_3$};
\node[circle,draw,inner sep=0pt,minimum size=12pt] (a) at (14,-0.5) {\scriptsize$v_4$};
\node[circle,draw,inner sep=0pt,minimum size=12pt] (a) at (14,-1) {\scriptsize$v_5$};
\node[circle,draw,inner sep=0pt,minimum size=12pt] (a) at (14,-1.5) {\scriptsize$v_6$};
\node[circle,draw,inner sep=0pt,minimum size=12pt] (a) at (14,-2) {\scriptsize$v_7$};
\node[circle,draw,inner sep=0pt,minimum size=12pt] (a) at (14,-2.5) {\scriptsize$v_8$};
\node[circle,draw,inner sep=0pt, minimum size=12pt] (a) at (14,-3) {\scriptsize$v_9$};
\node[circle,draw,inner sep=0pt,minimum size=12pt] (a) at (14,-3.5) {\scriptsize$v_{10}$};
\node[circle,draw,inner sep=0pt,minimum size=12pt] (a) at (14,-4) {\scriptsize$v_{11}$};
\node[circle,draw,inner sep=0pt,minimum size=12pt] (a) at (14,-4.5) {\scriptsize$v_{12}$};
\node[circle,draw,inner sep=0pt,minimum size=12pt] (a) at (14,-5) {\scriptsize$v_{13}$};
\draw[-] (11,1.4) rectangle (15.2,-5.5);

\draw [-{Latex[length=2.5mm]}] (15.2,-2) -- (18.5,-2);

\node[inner sep=0pt,align=center] (votestitle) at (23.1,1.7)
{\huge\textbf{Discriminative Model} $d$\\\\\huge CNN for Text classification};
\end{tikzpicture}